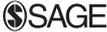

# Algorithmic failure as a humanities methodology: Machine learning's mispredictions identify rich cases for qualitative analysis

Jill Walker Rettberg

## Abstract
This commentary tests a methodology proposed by Munk et al. (2022) for using failed predictions in machine learning as a method to identify ambiguous and rich cases for qualitative analysis. Using a dataset describing actions performed by fictional characters interacting with machine vision technologies in 500 artworks, movies, novels and videogames, I trained a simple machine learning algorithm (using the kNN algorithm in R) to predict whether or not an action was active or passive using only information about the fictional characters. Predictable actions were generally unemotional and unambiguous activities where machine vision technologies were treated as simple tools. Unpredictable actions, that is, actions that the algorithm could not correctly predict, were more ambivalent and emotionally loaded, with more complex power relationships between characters and technologies. The results thus support Munk et al.'s theory that failed predictions can be productively used to identify rich cases for qualitative analysis. This test goes beyond simply replicating Munk et al.'s results by demonstrating that the method can be applied to a broader humanities domain, and that it does not require complex neural networks but can also work with a simpler machine learning algorithm. Further research is needed to develop an understanding of what kinds of data the method is useful for and which kinds of machine learning are most generative. To support this, the R code required to produce the results is included so the test can be replicated. The code can also be reused or adapted to test the method on other datasets.

## Keywords
Machine vision, machine learning, qualitative methodology, machine anthropology, digital humanities, algorithmic failure

## Introduction

In a recent paper, Munk, Olesen and Jacomy argue that it is the failed predictions of machine learning that are the most interesting for a qualitative researcher (2022). 'If the ambition is thick description and explication rather than formalist cultural analysis and explanation', Munk et al. argue, 'then a failure to predict is more interesting than accuracy'. Their study shows how cases where their neural network fails to predict which emoji a Facebook commenter would choose are in fact cases where the underlying data are much more subtle, ambiguous, and from an anthropological standpoint, interesting, than the cases where the algorithm makes a correct prediction. By identifying unpredictable cases, machine learning can thus support the fieldwork stage of ethnographers rather than attempting to identify laws and underlying structures.

I will call this aspect of Munk et al.'s method *algorithmic failure*. Algorithmic failure uses the *mispredictions* of machine learning to identify cases that are of interest for qualitative research.

Munk et al. situate their proposal in the tradition of anthropology, and especially Geertz's conception of cultural anthropology as 'an interpretative science in search of meaning rather than an explanatory one in search of law' (Munk et al., 2022; Geertz, 1973). Geertz saw this as an argument against his time's computational anthropology. This corresponds to a common criticism of the digital humanities: the results digital methods provide are often already known or they are reductive (Da, 2019). Munk

Department of Linguistic, Literary and Aesthetic Studies, University of Bergen, Bergen, Norway

**Corresponding author:**
Jill Walker Rettberg, Department of Linguistic, Literary and Aesthetic Studies, University of Bergen, Bergen 5020, Norway.
Email: jill.walker.rettberg@uib.no

et al.'s method sidesteps such critiques by using machine learning not to replace human interpretation, but to sort through vast amounts of data to find the most worthwhile cases for interpretation.

This commentary offers a test of algorithmic failure as a methodology by applying it to a dataset in a different field and by using a simpler machine learning algorithm. My results show that the method has potential for broader application. The R code and dataset used in this test are available so other scholars can further develop the method (Rettberg, 2022; Rettberg et al., 2022a).

## Dataset and methodology

The dataset I will test the method on is from the *Database of Machine Vision in Art, Games and Narratives*, which contains structured analyses of how machine vision technologies are represented and used in 500 digital artworks, video games, novels and movies. The dataset is fully documented in Rettberg et al. (2022b). The subset of data used for this test consists of 747 verbs describing an interaction between fictional characters and machine vision technologies. Verbs are not extracted from the works themselves but are interpretations assigned by the research team to describe the interactions, and they are either active ('the character is *scanning*') or passive ('the character is *scanned*'). Each verb is associated with information about the traits (gender, species, race/ethnicity, age and sexuality[1]) of the fictional characters involved in the machine vision situations.

For instance, when documenting a 'machine vision situation' in S. B. Divya's novel *Machinehood* (2021, p. 28), we described the protagonist Welga as using her ocular implants for *navigating*, *assessing* and *protecting*. The actions in this situation are all active, and the character Welga has the traits of adult, human, heterosexual, female and person of colour.

To test the algorithmic failure method, I decided to use machine learning to predict whether a verb was active or passive based only on the traits of the fictional characters that performed the verb. If the algorithmic failure method works, the actions the algorithm *can't* correctly predict will be the most interesting.

## Distribution of character traits compared to active or passive verbs

Preliminary data analysis had already shown some correlation between a character's traits and the actions they take when interacting with machine vision. As shown in Figure 1, adults are more likely than children to take an active role when interacting with machine vision technologies. White and Asian people are portrayed as taking more active roles than others. Robot characters are more active than humans, and fictional or animal-like species are less active.

Some categories appear very significant only because there are so few cases. For instance, the only three explicitly trans characters (all trans women) interacting with machine vision technologies in the 500 works in the dataset are all strong characters who take active roles when interacting with technology. This is certainly a striking finding worth qualitative analysis, but too small a sample to generalise from. Figure 2 shows the same distribution but using the actual count of times a verb is used.

## Training the algorithm

To test the value of using algorithmic failure as a method for identifying interesting cases, I organised the 747 verbs as a contingency table where each row shows a verb and the columns show how many characters with each trait use that verb. Using 70% of the dataset as training data, I ran a kNN algorithm that predicted whether each verb in the remaining 30% of the dataset was active or passive, using the method described by Lantz (2019).

kNN stands for *k* nearest neighbours, where *k* is the number of nearest neighbours the algorithm should compare each observation to in order to make its prediction. Trying various values of *k* showed that $k = 1$ had the highest accuracy for this dataset. That means that the algorithm was comparing each row in the test data to just one 'nearest neighbour' in the training data. Accuracy was only 56%, and the algorithm is more likely to predict active verbs correctly (64.6%) than passive verbs (35.8%). This might be terrible if our goal was an accurate prediction, but as I aim to show in the following, 'bad' machine learning can be quite adequate if the goal is to identify rich cases for analysis rather than to find general rules.

Although kNN is less sophisticated than Munk et al.'s neural network, it is also much simpler to code, faster to run and more energy-efficient, making it accessible to researchers with cheap computers and rudimentary skills in machine learning. This is a great advantage because it means humanities scholars can realistically hope to do their own coding.

## Predictable actions

Munk et al. argue that we don't need to understand *why* the model made false predictions. What is interesting is *which cases* it failed to predict. To test the method, we need to evaluate whether the faulty predictions are in fact the most interesting for qualitative analysis.

Eight of the 10 most frequently used actions in the test dataset were correctly predicted: *searching* (used in 43 interactions between characters and machine vision technologies), *analysed* (42), *analysing* (39), *classified* (38),

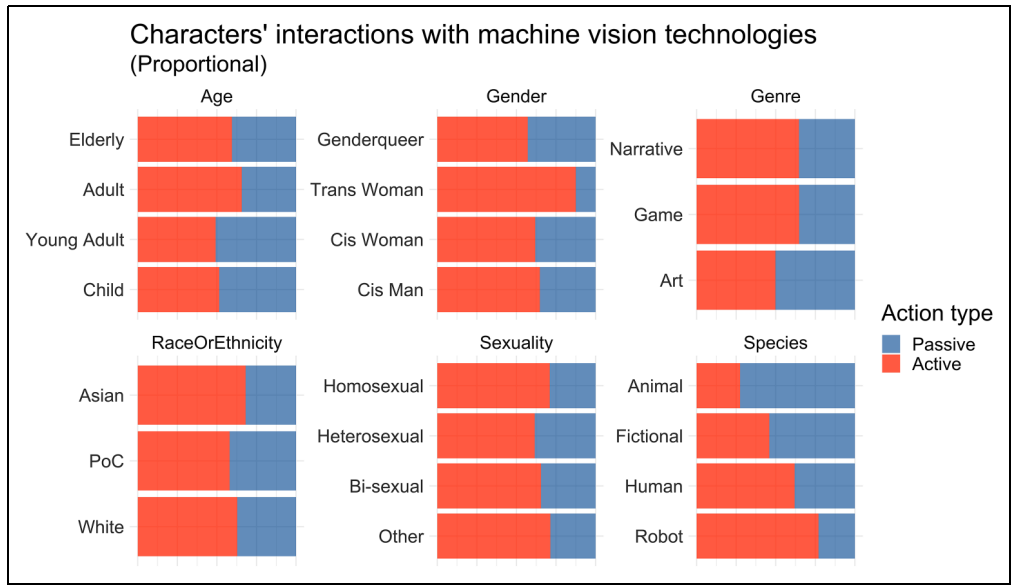

**Figure 1.** Distribution of active and passive stance to machine vision technology by character trait and genre. The dataset includes 747 unique verbs used 3439 times.

*hacking* (38), *investigating* (28), *helping* (25) and *hunted* (23). For the rest of the actions, accuracy is only 54%. This means that the algorithm is performing quite badly, but that doesn't necessarily matter if its mispredictions identify the most interesting cases.

As our research team was analysing works and entering data, we realised quite quickly that a lot of the actions were, we admitted to each other, rather boring. You would expect machine vision to be used for watching and analysing: that's the primary purpose of technologies like surveillance cameras and facial recognition algorithms. Predictably enough, a lot more male than female characters use machine vision for *killing*, just as you would expect given societal gender stereotypes. Whether or not you find such examples boring, it is clear that the algorithm did find something *predictable* about the association between these actions and the traits of the characters that were performing them. These are the sorts of results critics of digital humanities tend to criticise as being so obvious as to be a waste of effort.

## Most frequently used false passives

Interestingly, the false passives – that is actions the algorithm predicted were passive but that are actually active – describe far more ambiguous cases: interactions where the distribution of agency and power between the character and the technology is complicated and not immediately obvious. The most frequently used false passives are *learning* (27), *protecting* (19), *manipulating* (18), *attacking* (17), *spying* (17), *destroying* (11), *revealing* (11), *running* (9), *chasing* (8) and *explaining* (6).

Although the verbs are technically active, many seem to describe someone who is not in a position of power but reclaims agency. Someone who is *attacking* may be a powerful aggressor but may also be acting in self defence. Using machine vision technology for *manipulating* or *spying* might well be the work of an underdog.

One character in the dataset who is *attacking* is Amber Rose 348, a human-AI construct embedded in a spaceship in Wijeratne's novel *The Salvage Crew* (2020). Amber Rose overfits his image recognition algorithms because his crew is threatened. However, the overfitting makes the algorithms interpret random input as potential threats, causing Amber Rose to panic: he is *attacking* shadows. The action *attacking* is technically active, but in practice futile. This is certainly a case that could be generative in qualitative analysis.

## Most frequently used false actives

The 10 most frequently used actions that were actually passive but that the algorithm predicted to be active describe characters that are *scanned* (61), *exposed* (12), *projected* (12), *targeted* (11), *deceived* (9), *oppressed* (7), *assisted* (6), *copied* (6), *questioned* (6) and *alerted* (3).

Apart from the outlier *scanned*, which is the most frequently taken action in the whole test dataset, the false actives are interesting in that almost all imply an intense interaction with another agent. These characters are *exposed, targeted, deceived, oppressed and alerted*.

As we know, the algorithmic failure method doesn't provide explanations; it just offers us a set of interesting cases for further analysis. The next step would be to do a more qualitative analysis of these interactions, either by using data visualisations to analyse the actions of other agents these *exposed* and *targeted* characters are interacting with or by using more traditional methods of interpretation of the movies, games and other works, such as literary or semiotic analysis. My aim in this commentary is to demonstrate that the algorithmic failure method has promise, not to provide a thorough analysis of this dataset, so I will leave further analysis to other papers and conclude by situating the method in critical scholarship on machine learning.

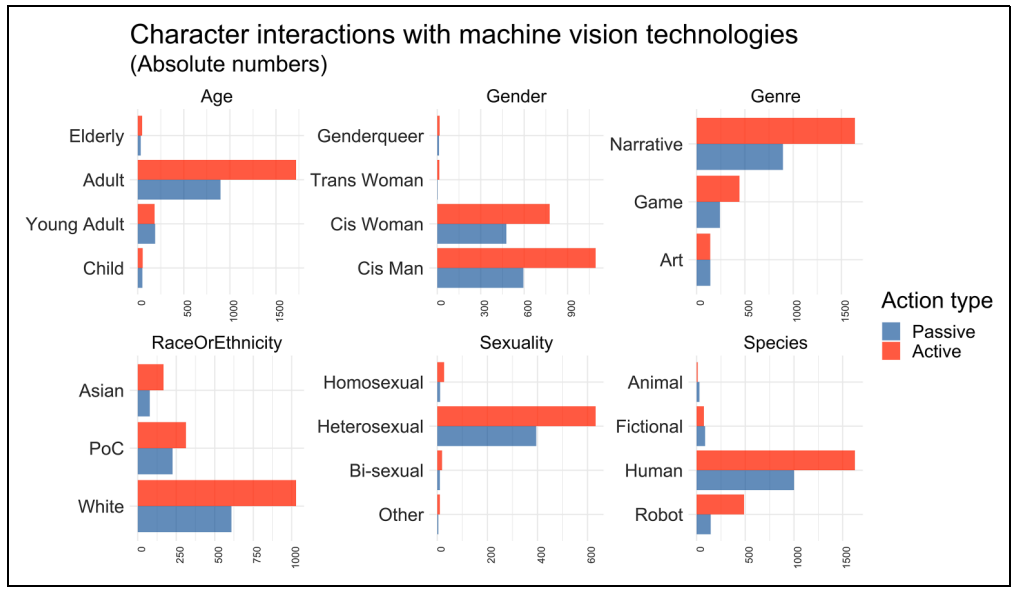

**Figure 2.** The same data as shown in Figure 1, but showing the absolute numbers rather than the proportions. The counts on the x-axis show the number of times characters with those particular traits engage in active or passive actions in machine vision situations in the 500 works we analysed.

### Using machine learning against the grain

Embracing algorithmic failure as a productive methodology upends the typical assumption that data science should lead to increased objectivity and accuracy. This assumption of data-driven objectivity is in fact one of the main tensions between qualitative methodologies and data science. There is a wealth of critical scholarship pointing out the problems of algorithmic bias (Chun 2022, Benjamin 2019) and critiquing the naive assumption that data could ever be treated as 'raw' rather than constructed (Gitelman 2013). Scholars have noted that data visualisations and models can express some kinds of knowledge well but render others invisible (Rettberg 2020) and can even function as self-fulfilling predictions or what Michelle Murphy (2017) calls *phantasmagrams*. Data analysis and data visualisation tend to obscure the 'troubling details' (Drucker 2014: 92) that are at the heart of qualitative research.

Algorithmic failure as a research strategy uses machine learning against the grain and has a lot in common with strategies used by artists. Artists exploring AI and machine vision technologies use failed predictions to showcase bias as in Paglen and Crawford's *ImageNet Roulette* (2019). Another example is Jake Elwes's work, which calls attention to the strangeness of the neural network outputs by emphasising them and 'queering the dataset' Elwes (2016).

The failed predictions of machines let us use machine learning as a collaborator, using algorithmic failures as 'technical intuitions (that) function as an interface between technical and human cognizers' (Kronman 2020). To some extent, this aligns with *human-in-the-loop* approaches to machine learning. Such approaches emphasise the collaboration and iterative cycle between human and machine, but their goal is generally to improve the algorithm's performance rather than to foster qualitative interpretation (Mosqueira-Rey et al. 2022).

As we work to develop machine learning methodologies that support the epistemologies of qualitative research, embracing algorithmic failure may be a productive way forward. The potential of failure has been noted by a number of scholars. Louise Amoore describes worries about failed predictions or that a model is over-fitting to the data as 'moments not of a lack or an error, but of a teeming plentitude of doubtfulness' (Amoore, 2020). The failed predictions in my study are far less dramatic than what Lauren Bridges calls the '*cybernetic rupture* where pre-existing biases and structural flaws make themselves known' (Bridges 2021: 13) but retain some of the celebration of failure that Halberstam describes in *The Queer Art of Failure* (Halberstam 2011).

## Next steps

Both in Munk, Olesen and Jacomy's study and in this test, failed predictions identified more ambiguous and generative cases than accurate predictions, with higher potential for productive qualitative analysis. Although the correctly predicted verbs are 'boringly' obvious, like the many cases of people using machine vision for *searching* or *analysing*, actions that the algorithm mispredicted suggest complex power dynamics between actors. This provides a rich and generative foundation for further qualitative research and data analysis.

Although algorithmic failure appears successful both in Munk et al.'s original study and my test of it, these successes could themselves be 'false positives'. Perhaps my data had too many variables and too little structure. Perhaps we should try more sophisticated machine learning techniques.

And yet, if the method identifies interesting cases, does it matter whether or not it is robust from a data science perspective? If we are to develop methods that genuinely combine data science with humanities research this is an epistemological question we will need to consider.

The code included with this brief commentary provides a simple method for reproducing my analysis and can be adapted to test the method on other datasets. Other types of machine learning algorithms should also be tested, especially considering the possible flaws of the kNN algorithm.

Machine learning was developed in quantitative fields such as statistics, mathematics and computer science. As the qualitative sciences work with larger datasets, we need to develop methodologies for using machine learning that build upon the epistemologies that are specific to the humanities and social sciences, and that support interpretation, uncertainty and detail. The algorithmic failure method has the potential to do just that.

## Acknowledgements

This research has received funding from the European Research Council (ERC) under the European Union's Horizon 2020 research and innovation programme (grant agreement No 771800).

## Data Availability Statement

The software used in the analysis is available on Github at https://github.com/jilltxt/algorithmicfailure and consists of a file with R scripts and comments explaining the code (Rettberg, 2022). The dataset is available on DataverseNO (Rettberg et al., 2022a), and there is also a data paper providing documentation (Rettberg et al. 2022b). *The Database of Machine Vision in Art, Games and Narratives* can also be viewed in a web browser at https://machine-vision.no.

## Declaration of conflicting interests

The author(s) declared no potential conflicts of interest with respect to the research, authorship, and/or publication of this article.

## Funding

The author(s) disclosed receipt of the following financial support for the research, authorship, and/or publication of this article: This work was supported by the H2020 European Research Council (grant number 771800).

## ORCID iD

Jill Walker Rettberg  https://orcid.org/0000-0003-2472-3812

## Note

1. Race, ethnicity, gender and sexuality are very difficult and sensitive categories to try to lock down in categorical data like this, but we wanted to include them because algorithmic bias is a major problem in society. For this analysis, 'PoC' includes a range of portrayed races and ethnicities, including Black people and indigenous people. «Genderqueer» includes characters that are represented as non-binary or where their gender is explicitly queered in other ways. The code provided (Rettberg 2022) demonstrates how to recode these categories, as terms are constantly evolving, and are also used differently in different geographic areas and cultural contexts. The ethics and implications of this are addressed briefly in Rettberg et.al. 2022b (p. 16) and will be addressed further in future work.

## References

Amoore L (2020) Doubt and the algorithm: On the partial accounts of machine learning. *Theory, Culture & Society* 36(6): 147–169.

Benjamin R (2019) *Race After Technology: Abolitionist Tools for the New Jim Code*. Cambridge, UK: Polity.

Bridges LE (2021) Digital failure: Unbecoming the "good" data subject through entropic, fugitive, and queer data. *Big Data & Society* 8(1). January 2021.

Chun WHK (2022) *Discriminating Data: Correlation, Neighborhoods, and the New Politics of Recognition*. Cambridge, MA: MIT Press.

Da NZ (2019) The computational case against computational literary studies. *Critical Inquiry* 45(3): 601–639.

Divya SB (2021) *Machinehood*. New York: Saga Press.

Drucker J (2014) *Graphesis: Visual Forms of Knowledge Production*. Cambridge, MA: Harvard University Press.

Elwes J (2016) Machine Learning Porn. https://www.jakeelwes.com/project-MLPorn.html

Gitelman L (ed.) (2013) *Raw Data Is an Oxymoron*. Cambridge MA: MIT Press.

Geertz C (1973) *The Interpretation of Cultures*. New York: Basic books.

Halberstam J (2011) *The Queer Art of Failure*. Durham: Duke UP.

Kronman L (2020) Intuition machines: Cognizers in complex human-technical assemblages. *A Peer-Reviewed Journal About* 9(1): 54–68.

Lantz B (2019) *Machine Learning with R: Expert Techniques for Predictive Modeling*. 3rd ed. Birmingham: Packt.

Mosqueira-Rey E., Hernández-Pereira E., Alonso-Ríos D., et al. (2022) Human-in-the-loop machine learning: A state of the art. *Artificial Intelligence Review*. doi: 10.1007/s10462-022-10246-w.

Munk AK, Olesen AG and Jacomy M (2022) The Thick Machine: Anthropological AI between explanation and explication. *Big Data & Society*, 9(1). doi: 10.1177/20539517211069891.

Murphy M (2017) *The Economization of Life*. Durham: Duke University Press Books.

Paglen T and Crawford K (2019) ImageNet Roulette. https://excavating.ai

Rettberg JW (2020) Ways of knowing with data visualizations. In: Engebretsen M and Kennedy H (eds) *Data Visualization in Society*. Amsterdam: Amsterdam University Press, pp. 35–47.

Rettberg JW (2022) Scripts for Testing Algorithmic Failure. Zenodo. doi: 10.5281/ZENODO.7075829.

Rettberg JW, Kronman L, Solberg R, et al. (2022a) A Dataset Documenting Representations of Machine Vision Technologies in Artworks, Games and Narratives'. DataverseNO. doi: 10.18710/2G0XKN.

Rettberg JW, Kronman L, Solberg R, et al. (2022b) Representations of machine vision technologies in artworks, games and narratives: Documentation of a dataset. *Data in Brief* 42. doi: 10.1016/j.dib.2022.108319.

Wijeratne Y (2020) *The Salvage Crew*. Fort Worth, TX: Aethon Books.